\documentclass[runningheads]{llncs}

\usepackage{booktabs}

\usepackage{graphicx}
\usepackage{amsmath}
\usepackage{amssymb}

\usepackage{tikz}
\usetikzlibrary{arrows.meta}

\usepackage{booktabs}

\usepackage{subcaption}
\usepackage{caption}
\usepackage{url}
\usepackage{hyperref}
\hypersetup{colorlinks={true},linkcolor={blue},citecolor=green}
\usepackage[nameinlink]{cleveref}
\Crefname{figure}{Figure}{Figures}
\Crefname{equation}{Equation}{Equations}
\Crefname{table}{Table}{Tables}
\Crefname{section}{Section}{Sections}

\newcommand{\Proc}{\mathsf{Proc}}
\newcommand{\Proj}{\mathsf{Proj}}
\newcommand{\Zon}{\mathsf{Zone}}
\newcommand{\ErrWrd}{\mathsf{ErrWord}}
\newcommand{\Err}{\mathsf{Err}}

\newcommand{\ProcObs}{\mathrm{Proc}^{Obs}}

\newcommand{\ErrWrdObs}{\mathrm{ErrWrd}^{{Obs}}}

\begin{document}
\title{Simple Root Cause Analysis by Separable Likelihoods}
%
%
\author{Maciej Skorski\inst{1}}
\authorrunning{M. Skorski}
%
\institute{DELL\\\email{maciej.skorski@gmail.com}}
\maketitle              
\begin{abstract}
Root Cause Analysis for Anomalies is challenging because of the trade-off between the accuracy and its explanatory friendliness, required for industrial applications. In this paper we propose a framework for simple and friendly RCA within the Bayesian regime under certain restrictions (that Hessian at the mode is diagonal, here referred to as \emph{separability}) imposed on the predictive posterior. We show that this assumption is satisfied for important base models, including Multinomal, Dirichlet-Multinomial and Naive Bayes. To demonstrate the usefulness of the framework, we embed it into the Bayesian Net and validate on web server error logs (real world data set).
\end{abstract}

\section{Introduction}

\subsection{Anomaly Detection and Root Cause Analysis}
In the likelihood-based approaches to anomaly detection, a generative probabilistic model for data is learned and used to evaluate new data records. Anomalies are defined as the records with unusually low likelihood. An example is the {Z-score} measure for 1-dimensional data, which fits the Gaussian distribution to the data (estimating the mean and variance) and scores observations in the decreasing order with respect to the likelihood; for its simplicity it is widely used in explanatory data analysis, quality controls and other industrial applications. The challenge with real data sets, however, is that they usually contains both continuous and categorical features, as well as inter dependencies (in particular anomaly scores cannot be applied independently). Interactions and dependencies can be effectively modeled by the modern framework of probabilistic graphical models~\cite{Koller:2009:PGM:1795555}. Further, simplicity can be traded for accuracy by using more sophisticated models as building blocks (for example more exotic base distributions or mixtures); only for multivariate counts several models have been proposed~\cite{Zhang2018}.

This paper concerns the constrained scenario of Root Cause Analysis (RCA) where in addition to identifying anomalies, a readable explanation (in terms of other features) is required. Because the purpose of RCA is to support business decision making, complexity and fit accuracy are often traded for explanatory abilities. This makes some powerful models (such as neural set) not adequate for this task~\cite{DBLP:journals/corr/SoleMRE17}. In this paper we show how to build, out of simple building blocks, an anomaly detection system for error logs. While our model is a fairly simple variant of Bayes Network,
the main added value is the proposed paradigm of \emph{determining anomaly contributions}, which is used to estimate how different features contribute to the likelihood of the anomaly data record. These scores can be used directly to perform efficient RCA which is illustrated by a case study on real data.

\subsection{Contribution}

\paragraph{Root-Cause Analysis for Separable Posteriors}

For the task of anomaly detection the main quantity of interest is the likelihood of the new data record $x=(x_1,\ldots,x_d)$ given the training data $\mathcal{D}$, called \emph{the predictive posterior}.
Assuming the generative process $p(\cdot |\theta)$ for the data, with some parameter $\theta$, the predictive posterior is given by
\begin{align*}
L(x) = p(x|\mathcal{D}) = \int p(x|\theta) p(\theta|\mathcal{D})\mbox{d}\theta.
\end{align*}
For the task of RCA it would be helpful to see how individual components $x_{i}$ impact the likelihood.
This is not possible in general, because posteriors are often not analytically tractable and only approximated by sampling.
However in certain cases the predictive posterior, after subtracting its mode, can be factorized into terms depending on individual terms $\theta_i$. More precisely, suppose that the predictive posterior log-likelihood can be written as
\begin{align}\label{eq:separable_posterior}
\log L(x) \approx  \log L(x^{*})  + \sum_{i} I(x_i).
\end{align}
where $x^{*} = \mathrm{argmax}_x \log L(x)$ is the mode. When the posterior obeys \Cref{eq:separable_posterior}  we say it is \emph{separable}. The term $I(x_i)$ can be then thought as \emph{influence} of the $i$-th coordinate of the data point $x$. Moreover, similarly to the notion of the averaged log-likelihood, these influences can be aggregated over several independent observations $x$ (e.g. at daily level).

This formula has the following intuitive meaning: we decompose \emph{the deficiency w.r.t. the mode} per individual dimensions; the deficiency is understood as the difference in the log-likelihood with respect to the mode and can be seen as a natural anomaly measure (note that $\sum_{i} I(x_i) \leqslant 0$ by the definition of $x^{*}$). We stress that it is important to subtract the mode in \Cref{eq:separable_posterior}, otherwise we explain the likelihood of a whole point, rather than its abnormal part.

It is worth mentioning that \Cref{eq:separable_posterior} can be characterized alternatively, by noticing that the hessian matrix $H$ satisfies
\begin{align}
\frac{\partial^2 H}{\partial x_i \partial x_j}(x^{*}) =[i=j]\cdot \frac{\partial I(x_i)}{\partial x_i}\cdot \frac{\partial I(x_j)}{\partial x_j}
\end{align}
hence is \emph{diagonal at the mode}.

We will show theoretical results on separability for two popular building blocks: the posterior of Dirichlet-Multinomial distribution and the posterior of categorical variable given category-dependent multivariate Bernoulli or Multinomial observations (for example, naive bayes text classification on the bag-of-words representation). They will be presented in \Cref{sec:separable_posteriors}; now we sketch a simpler example for illustration. Consider the multinomial model with total counts of $k$ and probability $p=(p_1,\ldots,p_d)$. The probability of counts $x=(k_1,\ldots,k_d)$ equals
\begin{align*}
L(x) = \binom{k}{k_1,\ldots,k_d}\prod_{i=1}^{d}p_i^{k_i}
\end{align*}
Denote by $q_i = \frac{k_i}{k}$ the observed frequencies. The log-likelihood normalized by the number of observations
can be approximated by Stirling formulas~\cite{DBLP:journals/corr/Shlens14c} establishing the connection to the \emph{Kullback-Leibler divergence of observed and real frequencies}, respectively $q=(q_1,\ldots,q_d)$ and $p=(p_1,\ldots,p_d)$.
\begin{align*}
\frac{1}{k}\log L(x) &\approx O(k^{-1}\log k) + D_{\mathrm{KL}}(q||p)\\
& = O(k^{-1}\log k) - \sum_{i=1}^{d}q_i\log \frac{q_i}{p_i}
\end{align*}
It is not hard to see that the logarithm of the mode for the multinomial distribution equals $O(\log k)$. Thus we obtain~\Cref{eq:separable_posterior} with $I(x_i)=q_i\log\frac{q_i}{p_i}$.

\paragraph{Case Study on Real Data}

We apply our framework to the real data set of error logs from company servers. Each record contains the number of errors for a given zone, project, procedure and the error message. The data was collected for more than 120 consecutive days.
A sample of the data set is shown in \Cref{tab:data}.
\begin{table}
\captionsetup{font=scriptsize}
\resizebox{\textwidth}{!}{
\begin{tabular}{|llllllr|}
\toprule
row\_id &     date & region &                    project\_name &                                     procedure\_name &                                       error\_detail &  err\_cnt \\
\midrule
15362 & 2018-04-01 &   EMEA &           GLOBAL\_ONLINE\_SERVICE &                                     EXPLODE\_BUNDLE &  Object reference not set to an instance of an ... &        3 \\
29308 & 2018-04-01 &   EMEA &                  YOJEG\_API &  YOJEG.Controllers.Configurator.Global.Glo... &                VerifyError:Invalid option selected &        1 \\
29222 & 2018-04-01 &   EMEA &  GDAS Services: CustomerService &                                                NaN &                          Operation: GetSalesPerson &       26 \\
3157  & 2018-04-01 &   EMEA &  GDAS Services: CustomerService &                                                NaN &  Operation: GetCustomer Exception: GDAS.Ex... &       77 \\
7801  & 2018-04-01 &   EMEA &                  YOJEG\_API &  YOJEG.Controllers.Configurator.Global.Glo... &           BuildError:InvalidOrderCodeOrCustomerSet &        5 \\
\bottomrule
\end{tabular}
}
\caption{Dataset for log errors.}
\label{tab:data}
\end{table}
\\The results will be discussed in \Cref{sec:rca}.

\subsection{Organization}

In \Cref{sec:separable_posteriors} we derive theoretical results for some separable posteriors.
In \Cref{sec:rca} we demonstrate our framework on the real-world data. The paper is concluded in \Cref{sec:conc}.

\section{Separable Posteriors}\label{sec:separable_posteriors}

\subsection{Dirichlet-Multinomial Model}

The Dirichlet-Multinomial Model (DM) is popular for modeling multivariate counts. As opposed to the plain multinomial model,
it models uncertainty in the probability parameter, which helps avoiding over-dispersion. 
\begin{align}\label{eq:DM_model}
\begin{aligned}
(p_1,\ldots,p_d) & \sim \mathsf{Dir}(\alpha_1,\ldots,\alpha_d) \\
(k_1,\ldots,k_d) &\sim \mathsf{Mult}(p_1,\ldots,p_d | k)
\end{aligned}
\end{align}
This model is analytically tractable, we utilize formulas derived in~\cite{Tu2014}.
\begin{align}
P((k_i)_i|\mathcal{D}) = \frac{\Gamma(k+1)}{\prod_i \Gamma(k_i+1)}\cdot \frac{\Gamma(\alpha')}{\prod_i \Gamma(\alpha'_i)}\cdot \frac{\prod_i \Gamma(k_i+\alpha'_i)}{\Gamma(k+\alpha')}
\end{align}
Where $\alpha'_i = \alpha_i + \sum_{x\in \mathcal{D}}\sum x^{i}$ and $\alpha'=\sum_{i}\alpha'_i$ or the sake of concise notation. By using the Stirling approximation we obtain
\begin{multline}
\log L  \approx \\ -k\sum_i\frac{k_i}{k}\log\frac{k_i}{k}-\alpha'\sum_i\frac{\alpha'_i}{\alpha'}\log\frac{\alpha'_i}{\alpha'} 
 +(k+\alpha')\sum_{i}\frac{k_i+\alpha'_i}{k+\alpha'}\log \frac{k_i+\alpha'_i}{k+\alpha'}
\end{multline}
In order to see separability we will apply the well known trick called \emph{Laplace approximation}, which is merely a multivariate Gaussian approximation to the predictive posterior (see for example~\cite{2017arXiv171108911D} for theoretical justifications). Technically, we expand the log-likelihood in a Talyor series around its mode, so that
linear term disappear (by the first-derivative test, as the mode maximizes the likelihood!) and quadratic terms correspond to 
the Gaussian terms. In our case, the second-order terms turn out to be diagonal hence we obtain separability.

In order to find the mode we need to use the Lagrangian because of the implicit constraint $k=\sum_{i}k_i$.
For some constant $C$, the mode satisfies\footnote{We extend the likelihood over non-integer frequencies as the gamma function is well-defined and the Stirling approximation works.}
\begin{align}\label{eq:mode_first_order}
-\log k^{\mathrm{MAP}}_i+\log (k^{\mathrm{MAP}}_i + \alpha'_i) + C = 0
\end{align}
which implies
\begin{align}
k^{\mathrm{MAP}}_i = \frac{k}{\alpha'}\cdot \alpha'_i.
\end{align}
By the Taylor expansion around the mode we obtain (note that the linear part disappears and the coefficients of the quadratic part are determined from the first order conditions \Cref{eq:mode_first_order})
\begin{align}
\log L((k_i)) & \approx \log L((k^{\mathrm{MAP}}_i)) - \frac{1}{2}\sum_i\frac{\alpha'_i}{(k^{\textrm{MAP}}_i+\alpha'_i)k^{\textrm{MAP}}_i}
\left(k_i-\frac{k}{\alpha'}\cdot \alpha'_i\right)^2 \nonumber \\
& = \log L((k^{\mathrm{MAP}}_i)) - \frac{1}{2}\sum_i\frac{1}{1+\frac{k}{\alpha'}}
\cdot  \frac{\left(k_i-\frac{k}{\alpha'}\cdot \alpha'_i\right)^2}{\frac{k}{\alpha'}\cdot \alpha'_i}
\end{align}
in the alternative notation $q_i =\frac{k_i}{k}$ (observed frequency) and $p^{\alpha'} = \frac{\alpha_i}{\alpha}$ (mode frequency) we have
\begin{lemma}[Predictive Posterior vs Mode for DM]
\begin{align}
\log L( q ) \approx \log L( p^{\mathrm{MAP}} ) - \frac{1}{2}\cdot \frac{\alpha'}{\alpha'+k}\cdot k\sum_i \frac{\left(q^{}_i-p^{\alpha'}_i \right)^2}{p^{\alpha'}_i}
\end{align}
\end{lemma}
Since usually $k\ll \alpha'$ ($\alpha'$ collects all occurrences over the training data) we have $\frac{\alpha'}{k+\alpha'}\approx 1$ and we conclude
\begin{corollary}[DM Posterior Predictive Impacts]\label{cor:dm_rca}
For the DM-model the impact for the $i$-th component in \Cref{eq:separable_posterior} equals
\begin{align}
I(k_i) \approx \frac{1}{2}\cdot k\sum_i \frac{\left(q^{}_i-p^{\alpha'}_i \right)^2}{p^{\alpha'}_i}
\end{align}.
\end{corollary}

\begin{remark}[Intuition]
The major reason for impacts being large negative is a significant relative increase in frequencies (observed vs posterior), under large volume. Indeed, let $q^{}_i=(1+r_i)p^{\alpha'}_i$ then the $i$-th impact equals $I(k_i) =r_i^2 p^{\alpha'}_i$.
\end{remark}

\subsection{BNB Model}

We prove separability only for Bernulli Naive Bayes (BNB) as we will be using this model in our case study. However, separability is not limited to the Bernoulli variant and can be also proved for Multionomial Naive Bayes.

The BNB model is popular for classification of short text messages. Texts are represented as as the $|V|$-dimensional boolean vectors where $V$ is the vocabulary. Each entry is a boolean number indicating occurrence of the word $w$ in a given text $\mathbf{w}$; we will use the notation $I(w\in \textbf{w})$. The model with Beta prior (which smooths zero-frequencies assuming extra "pseudocounts" of one for each class-word) can be written as
\begin{align*}
\begin{aligned}[rl]
 \forall c\in\mathcal{C}\forall w\in V&\quad p_{w|c} \sim \mathsf{Beta}(1,1)  \\
\forall c\in\mathcal{C}\forall w\in V &\quad I(w\in\textbf{w}|c) \sim \mathsf{Ber}(p_{w|c})
\end{aligned}
\end{align*}
where $\mathcal{C}$ is the set of classes (categories). Let $p_{w|c}$ and $p_c$ be posterior probabilities for word given class and class (estimated from the data). Then we have
\begin{proposition}[Predicitve Posterior for BNB]
Probability of the class $c$ given the vector of words $\textbf{w}\in\mathbb{R}^{V}$ is given by
\begin{align}\label{eq:BNB}
L(c|\textbf{w}) \propto p_c \cdot \prod_{w\in \textbf{w}} p_{w|c}^{I(w\in \textbf{w})}(1-p_{w|c})^{I(w\not\in V)}
\end{align}
where the proportionality constant is independent on $c$ (but depends on $\textbf{w}$).
\end{proposition}
By taking the logarithm of \Cref{eq:BNB} evaluated at $c$ and $c^{*}$ and subtracting (the unknown constant cancels) we obtain
\begin{lemma}[Predictive Posterior vs Mode for BNB]\label{lemma:bnb_impact}
For the Bernoulli Naive Bayes model, let $c^{*}$ be the most likely class given the sequence of words $\textbf{w}\subset V$. We have
\begin{multline}
\log L(c|\textbf{w})-\log L(c^{*}|\textbf{w}) =\\ \log \frac{p_c}{p_{c^{*}}\,} +\sum_{w\in \textbf{w}}\left[ I(w\in \textbf{w})\log\frac{p_{w|c}}{p_{w|c^{*}}}+I(w\not\in \textbf{w})\log\frac{1-p_{w|c}\,}{1-p_{w|c^{*}}}\right]
\end{multline}
\end{lemma}
From \label{lemma:bnb_impact} we immediately obtain the word impact.

\begin{corollary}[BNB Posterior Predictive Impact]\label{corr:word_daily_impact}
For the BNB-model the impact for the $w$-th word in \Cref{eq:separable_posterior} equals
\begin{align}
I(w) =\sum_{w\in \textbf{w}}\left[ I(w\in \textbf{w})\log\frac{p_{w|c}}{p_{w|{c^{}}^{*}}}+I(w\not\in \textbf{w})\log\frac{1-p_{w|c}\,}{1-p_{w|{c}^{*}}}\right]
\end{align}
where $c$ is the actual class.
\end{corollary}

\begin{remark}[Intuition]
The major reason for impacts $I(w)$ to be large negative is the presence of class-untypical words (so that $p_{w|c} \ll p_{w|c^*}$). The effect is stronger with large volume when evaluating averaged likelihoods.
\end{remark}

\section{Root Cause Analysis of Anomalies}\label{sec:rca}

\subsection{Generative Model}

\begin{figure}
\centering
\begin{tikzpicture}
\node (Zon) at (0,0) {Zone};
\node (Proj) at (-3,-3) {Project};
\node (Proc) at (0,-3) {Procedure};
\node (Err) at (3,-3) {Error};
\path [->] (Zon) edge node {} (Proj);
\path [->] (Zon) edge node {} (Proc);
\path [->] (Zon) edge node {} (Err);
\path [->] (Proj) edge node {} (Proc);
\path [->] (Proc) edge node {} (Err);
\end{tikzpicture}
\caption{TAN model for occurrences of a \emph{single error}.}
\label{fig:model}
\end{figure}
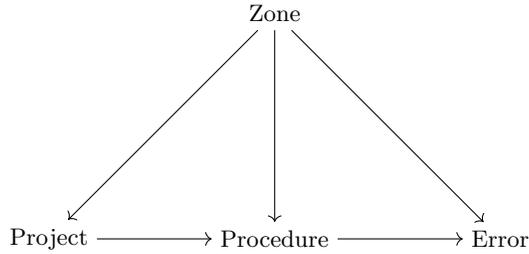

Before we apply the results of the previous section, we need to construct the joint model for all features in our data set.
We model the Data by a Bayes Net illustrated in \Cref{fig:model}. 
Every feature is dependent on zone (justification: different zones use servers in different location)
and at most one other feature (in the natural hierarchical way). Thus, the model is actually a Tree-Augmented Network (TAN).
These models generally allow for a feature-root relation and one more level of interaction. 
While TANs can capture non-trivial dependencies, they are computationally attractive since every node has at most two parents which reduces the size of internal conditional probability tables~\cite{Harini2015}.

More precisely, we assume 
\begin{align}
\begin{aligned}
\Proj|\Zon \sim \mathsf{Cat}(p=p(\Zon)) \\
\Proc|\Proj,\Zon \sim \mathsf{Cat}(p=p(\Proj,\Zon)) \\
\Err | \Proc,\Proj,\Zon \sim \mathsf{Ber}(p=p(\Proc,\Zon))
\end{aligned}
\end{align}
with empirical Dirichlet priors (estimated from data) for $\Proj,\Proc$ and non-informative Beta prior for $\Err$). Bernoulli distributions are over the (binarized) bag-of-word text representation of $\Err$.

Given the graph, the likelihood factorizes into likelihoods of individual features given parents; these models can be fit separately~\cite{Harini2015}. In our case
\begin{align*}
\Pr[\Proj,\Proc,\Err|\Zon] = \Pr[\Err|\Proc,\Zon]\cdot\Pr[\Proc|\Proj,\Err,\Zon] \cdot \Pr[\Proj|\Zon]
\end{align*}
We also use this fact to structure our anomaly detection: we will analyze separately anomalies in $\Proj,\Zon$ and separately in tuples $\Err,\Proc,\Zon$.
Since we are interested in discovering and explaining anomalies on the daily bases, we perform the inference day by day, training the algorithm on the past data. The model was implemented under Python package PyMC3~\cite{10.7717/peerj-cs.55}.

\subsection{RCA for Projects}

\begin{figure}
\centering
\begin{subfigure}{.45\textwidth}
\includegraphics[width=0.95\linewidth]{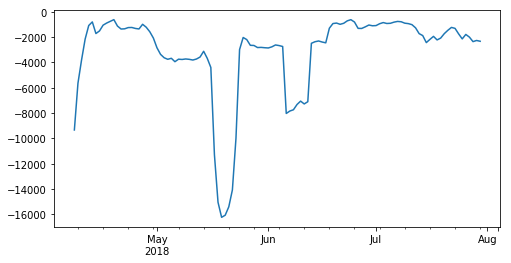}
\caption{Likelihood for $\Proj$, $\Zon=\mathrm{EMEA}$}
\end{subfigure}
\begin{subfigure}{.45\textwidth}
\includegraphics[width=0.95\linewidth]{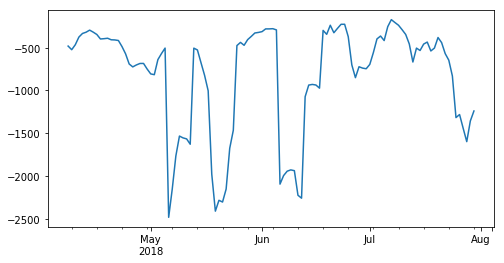}
\caption{Likelihood for $\Proj$, $\Zon=\mathrm{APJ}$}
\end{subfigure}
\caption{Project likelihoods by zone.}
\label{fig:project_like}
\end{figure}

The posterior for $\Proj$ given observed projects counts is Dirichlet-Multinomial. The daily-averaged likelihood is illustrated in
\Cref{fig:project_like}.

\paragraph{Anomalies 2018/05/17 and 2018/06/11, EMEA}

\begin{figure}
\centering
\begin{subfigure}{.49\textwidth}
\includegraphics[width=0.95\linewidth]{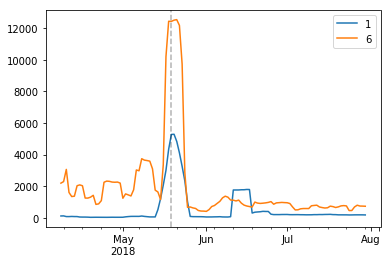}
\end{subfigure}
\begin{subfigure}{.49\textwidth}
\includegraphics[width=0.95\linewidth]{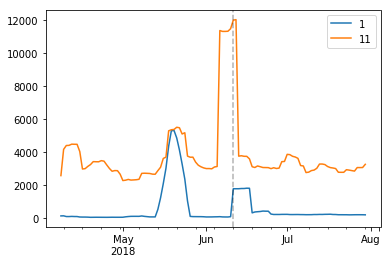}
\end{subfigure}
\caption{Daily hits by project ($\Zon=\mathrm{EMEA}$).}
\label{fig:project_anom_emea}
\end{figure}

By applying \Cref{cor:dm_rca} we obtain most impacting projects. We see that the anomalies corresponds to peaks in project hits as illustrated in \Cref{fig:project_anom_emea}.

\begin{figure}
\centering
\begin{subfigure}{.49\textwidth}
\includegraphics[width=0.95\linewidth]{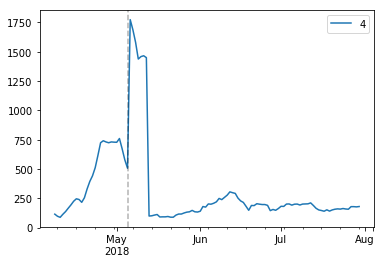}
\end{subfigure}
\begin{subfigure}{.49\textwidth}
\includegraphics[width=0.95\linewidth]{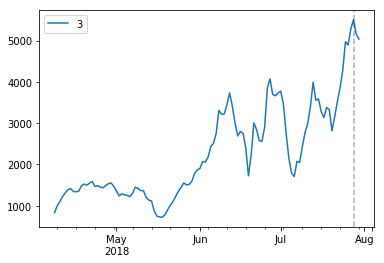}
\end{subfigure}
\caption{Daily hits by project ($\Zon=\mathrm{APJ}$).}
\label{fig:project_anom_apj}
\end{figure}

\paragraph{Anomalies 2018/05/07  and 2018/07/28, APJ}
By applying \Cref{cor:dm_rca} we obtain most impacting projects (we pick two). The anomalies again corresponds to peaks in project hits as illustrated in \Cref{fig:project_anom_apj}.

\subsection{RCA for Procedures and Error Messages}
According to our model, the distribution of procedures given error descriptions follows the classification Bernoulli Naive Bayes (BNB) model (where $\Proc$ is the class and $\Err$ is text; class priors are determined by fitting $\Proc[\Proc|\Proj,\Zon]$). To detect anomalies in errors, we evaluate \emph{how error messages impact procedures} rather than investigating $\label{eq:err_on_proc_zon}$ for individual errors.

\begin{figure}
\centering
\begin{subfigure}{.5\textwidth}
  \centering
  \includegraphics[width=1\linewidth]{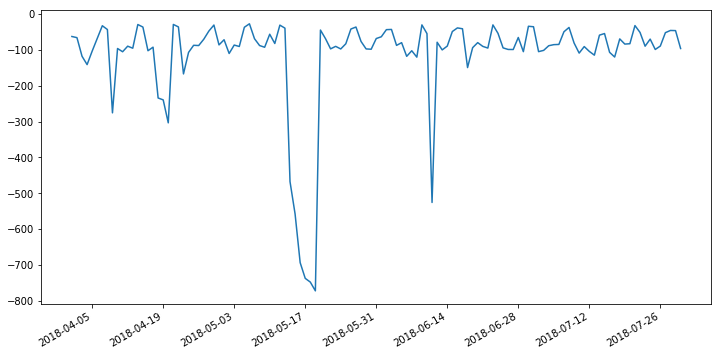}
  \caption{Likelihood of $\Proc$ in $\Zon=\text{EMEA}$}
  \label{fig:bnb_emea}
\end{subfigure}%
\begin{subfigure}{.5\textwidth}
  \centering
  \includegraphics[width=1\linewidth]{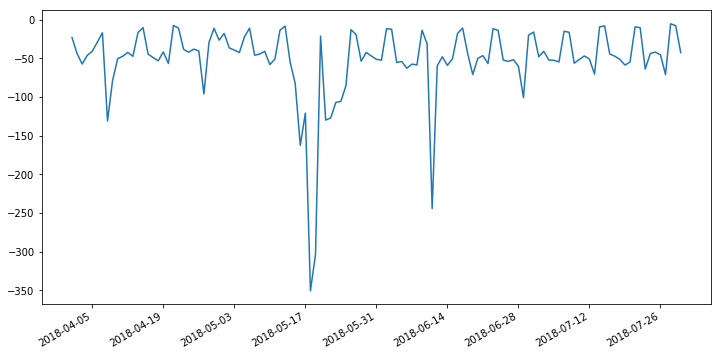}
  \caption{Likelihood of $\Proc$ in $\Zon=\text{APJ}$}
  \label{fig:bnb_emea}
\end{subfigure}
\caption{Likelihood of $\Proc$ split by $\Zon$.}
\label{fig:bnb}
\end{figure}

To detect anomalies on the daily level, we compute the daily-averaged likelihood and illustrate in \Cref{fig:bnb}

\paragraph{Anomaly 2018/05/17 in EMEA}

\begin{figure}
\centering
\includegraphics[width=0.6\linewidth]{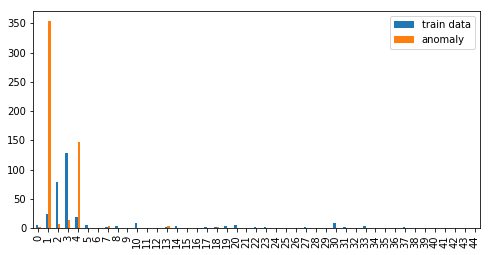}
\caption{Average daily hits of the keywords $S$ split by class ($\mathsf{Proc}$), for EMEA zone.}
\label{fig:BNB_emea_anom_0517}
\end{figure}

By \Cref{corr:word_daily_impact} we identify the set $$S=\{\text{'object', 'set', 'reference', 'instance', 'connection'}\}$$ of 3 keywords with biggest negative influence on the likelihood. By inspecting hits on these keywords (by hit we understand every message matching at least one word in $S$) across the classes we notice a huge difference between the anomaly day and the reference data set (see~\Cref{fig:BNB_emea_anom_0517}).

By inspecting message texts we also recognize the specific messages related to the keywords $S$. The result is summarized in \Cref{tab:BNB_anom1}.
\begin{table}
\captionsetup{font=scriptsize}
\centering
\begin{tabular}{|l|l|}
\hline
procedure & error message \\
\hline
$Proc_{1}$  & Object reference not set to an instance of an object \\
\hline
$Proc_{4}$ & Object reference not set to an instance of an object \\
\hline
\end{tabular}
\caption{RCA for anomaly 2018/05/17 EMEA.}
\label{tab:BNB_anom1}
\end{table}

\paragraph{Anomaly 2018/06/11 in EMEA}

\begin{figure}
\centering
\includegraphics[width=0.6\linewidth]{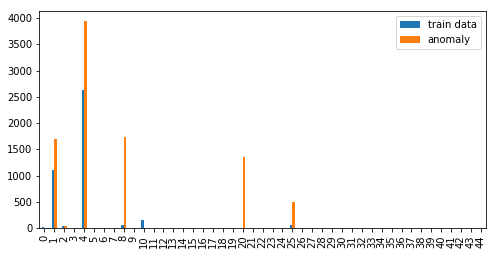}
\caption{Average daily hits of the keywords {'channel', 'timed', 'remote', 'returned', 'request'} split by class ($\mathsf{Proc}$) for EMEA zone.}
\label{fig:BNB_emea_anom_0611}
\end{figure}

By \Cref{corr:word_daily_impact} we identify the set $$S=\{\text{'channel', 'timed', 'remote', 'returned', 'request'}\}$$ of 5 keywords with biggest negative influence on the likelihood. By inspecting hits on these keywords across the classes we notice a significant shift between the anomaly day and the reference data set (see~\Cref{fig:BNB_emea_anom_0611}).

Having localized the keywords, we easily find procedures with biggest shifts and also the messages. The explanation is summarized in \Cref{tab:BNB_anom2}.
\begin{table}
\captionsetup{font=scriptsize}
\centering
\begin{tabular}{|l|l|}
\hline
procedure & error message \\
\hline
$Proc_{8}$  & The operation has timed out \\
\hline
$Proc_{25}$ & The request channel timed out \\
\hline
$Proc_{20}$ & The request failed with HTTP status 404 \\
\hline
\end{tabular}
\caption{RCA for anomaly 2018/06/11 EMEA.}
\label{tab:BNB_anom2}
\end{table}

\paragraph{Anomaly 2018/05/18 APJ}

\begin{figure}
\centering
\includegraphics[width=0.6\linewidth]{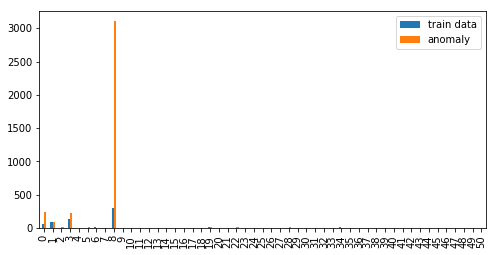}
\caption{Average daily hits of the keywords {'null', 'reference', 'set'} split by class ($\mathsf{Proc}$), for APJ zone.}
\label{fig:BNB_apj_anom_0518}
\end{figure}

By \Cref{corr:word_daily_impact} we identify the set $$S=\{\text{'null', 'reference', 'set'}\}$$ of 3 keywords with biggest negative influence on the likelihood. By inspecting hits on these keywords across the classes we notice a significant shift between the anomaly day and the reference data set (see~\Cref{fig:BNB_apj_anom_0518}).

The explanation by procedures and error messages is shown in \Cref{tab:BNB_anom3} below.
\begin{table}
\captionsetup{font=scriptsize}
\centering
\begin{tabular}{|l|l|}
\hline
procedure & error message \\
\hline
$Proc_{8}$  & argument is null \\
\hline
\end{tabular}
\caption{RCA for anomaly 2018/05/18 APJ.}
\label{tab:BNB_anom3}
\end{table}

\paragraph{Anomaly 2018/06/11 APJ}

\begin{figure}
\centering
\includegraphics[width=0.6\linewidth]{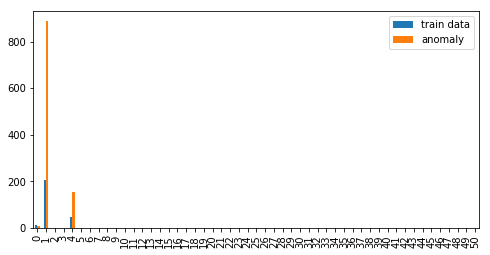}
\caption{Average daily hits of the keywords 'contract', 'gdas', 'contracts','target','invocation' split by class ($\mathsf{Proc}$), for APJ zone.}
\label{fig:BNB_apj_anom_0611}
\end{figure}

By \Cref{corr:word_daily_impact} we identify the set $$S=\{\text{'contract', 'gdas', 'contracts','target','invocation'}  \}$$ of 5 keywords with biggest negative influence on the likelihood. By inspecting hits on these keywords across the classes we notice a significant shift between the anomaly day and the reference data set (see~\Cref{fig:BNB_apj_anom_0611}).

The explanation by procedures and error messages is shown in \Cref{tab:BNB_anom4} below.
\begin{table}
\captionsetup{font=scriptsize}
\centering
\begin{tabular}{|l|l|}
\hline
procedure & error message \\
\hline
$Proc_{1}$  & Operation: GDAS.Exceptions.CustomerNotFoundException \\
\hline
$Proc_{4}$ & Exception has been thrown by the target of an invocation  \\
\hline
\end{tabular}
\caption{RCA for anomaly 2018/06/11 APJ.}
\label{tab:BNB_anom4}
\end{table}

\section{Conclusion}\label{sec:conc}

We proposed a framework for anomaly detection and root cause analysis based on \emph{separable posterior approximation}.
This approximation has been proved for the case of Multionomial, Dirchlet-Multinomal and Naive Bayes Models.
The validation on the real data set shows that the framework detects anomalies and offers reasonable and simple explanations.

\bibliographystyle{amsalpha}
\bibliography{citations}

\end{document}